\documentclass[]{gtech}
\usepackage{amssymb}
\usepackage{multirow}
\usepackage{bigdelim}
\usepackage{todonotes}
\usepackage{longtable}
\usepackage{tabularray}
\usepackage{wrapfig}
\usepackage[most]{tcolorbox} 
\usepackage{xcolor}
\usepackage{url}
\usepackage{hyperref}
\usepackage{enumitem}
\usepackage{graphicx}
\usepackage{array}
\usepackage{booktabs}
\usepackage{tabularx}
\usepackage{siunitx}
\usepackage{makecell}  
\usepackage{multicol} 

\vspace{-5cm}
\title{
    \raisebox{-0.4ex}{\includegraphics[height=25pt]{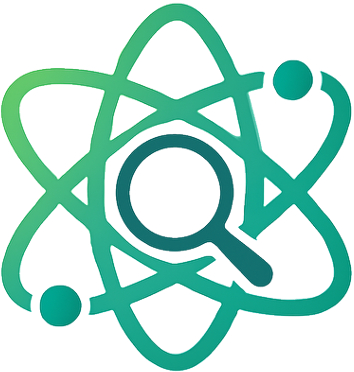}}
    Atom-Searcher: Enhancing Agentic Deep Research via Fine-Grained Atomic Thought Reward
}
\author[*]{Yong Deng}
\author[*]{Guoqing Wang}
\author[*]{Zhenzhe Ying}
\author[*]{Xiaofeng Wu}
\author{Jinzhen Lin, Wenwen Xiong, Yuqin Dai, Shuo Yang, Zhanwei Zhang, Qiwen Wang, Yang Qin, Yuan Wang, Quanxing Zha, Sunhao Dai, Changhua Meng}
\contribution[*]{Core Contributors}

\affiliation{Ant Group}

\vspace{-20pt}
\abstract{
Large language models (LLMs) exhibit remarkable problem-solving abilities, but struggle with complex tasks due to static internal knowledge. Retrieval-Augmented Generation (RAG) enhances access to external information, yet remains limited in multi-hop reasoning and strategic search due to rigid workflows. Recent advancements in agentic deep research empower LLMs to autonomously reason, search, and synthesize information. However, current approaches relying on outcome-based reinforcement learning (RL) face critical issues such as conflicting gradients and reward sparsity, limiting performance gains and training efficiency. To address these, we first propose Atomic Thought, a novel LLM thinking paradigm that decomposes reasoning into fine-grained functional units. These units are supervised by Reasoning Reward Models (RRMs), which provide Atomic Thought Rewards (ATR) for fine-grained guidance. Building on this, we propose Atom-Searcher, a novel RL framework for agentic deep research that integrates Atomic Thought and ATR. Atom-Searcher uses a curriculum-inspired reward schedule, prioritizing process-level ATR early and transitioning to outcome rewards, accelerating convergence on effective reasoning paths. Experiments on seven benchmarks show consistent improvements over the state-of-the-art. Key advantages include: (1) Atom-Searcher scales computation at test-time. (2) Atomic Thought provides supervision anchors for RRMs, bridging deep research tasks and RRMs. (3) Atom-Searcher exhibits more interpretable, human-like reasoning patterns.
}

\gtechdata[Code]{\url{https://github.com/antgroup/Research-Venus}}

\begin{document}
\maketitle
\vspace{-0.25cm}
\begin{center}

\includegraphics[width=1\textwidth]{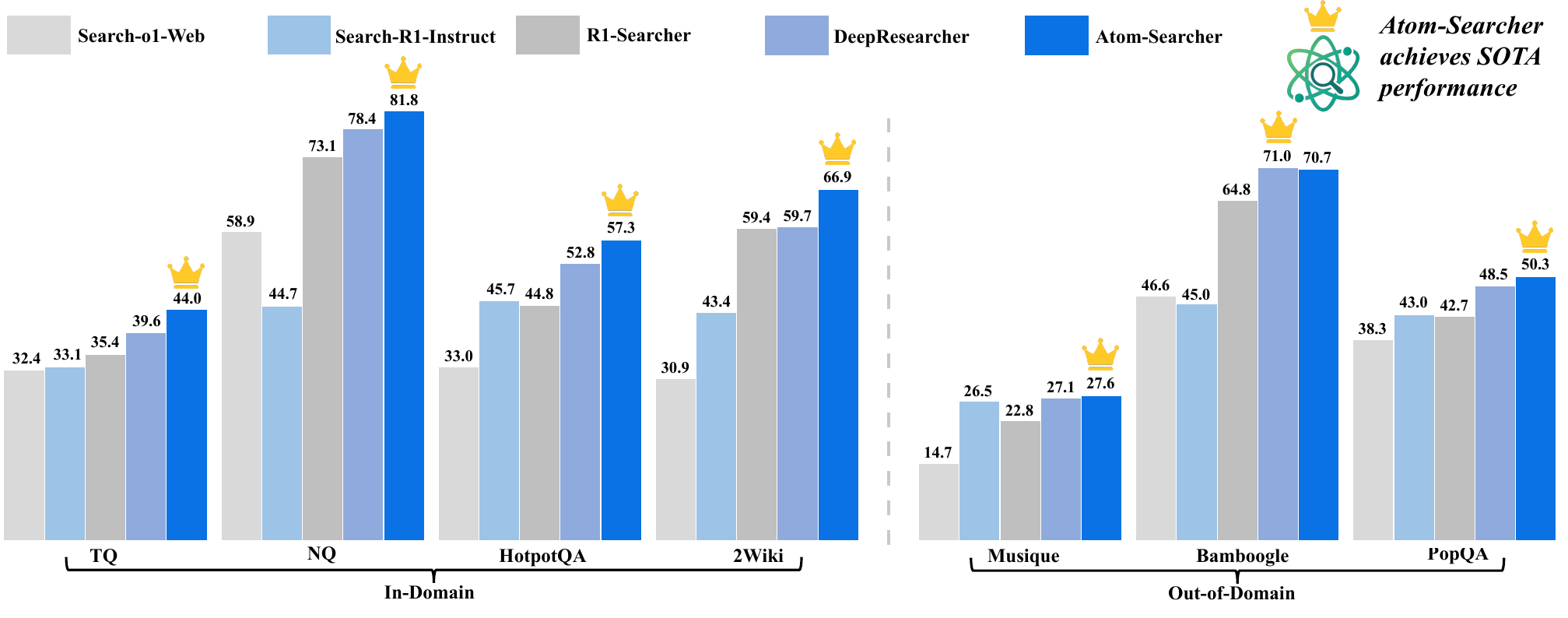}
\vspace{-7mm}
\captionof{figure}{Atom-Searcher achieves SOTA performance on both in-domain and out-of-domain benchmarks.}
\label{fig:exp}
\vspace{-5mm}
\end{center}

\section{Introduction}
\begin{wrapfigure}{r}{0.46\textwidth}
  \vspace{-12pt}
  \centering
  \includegraphics[width=1\linewidth]{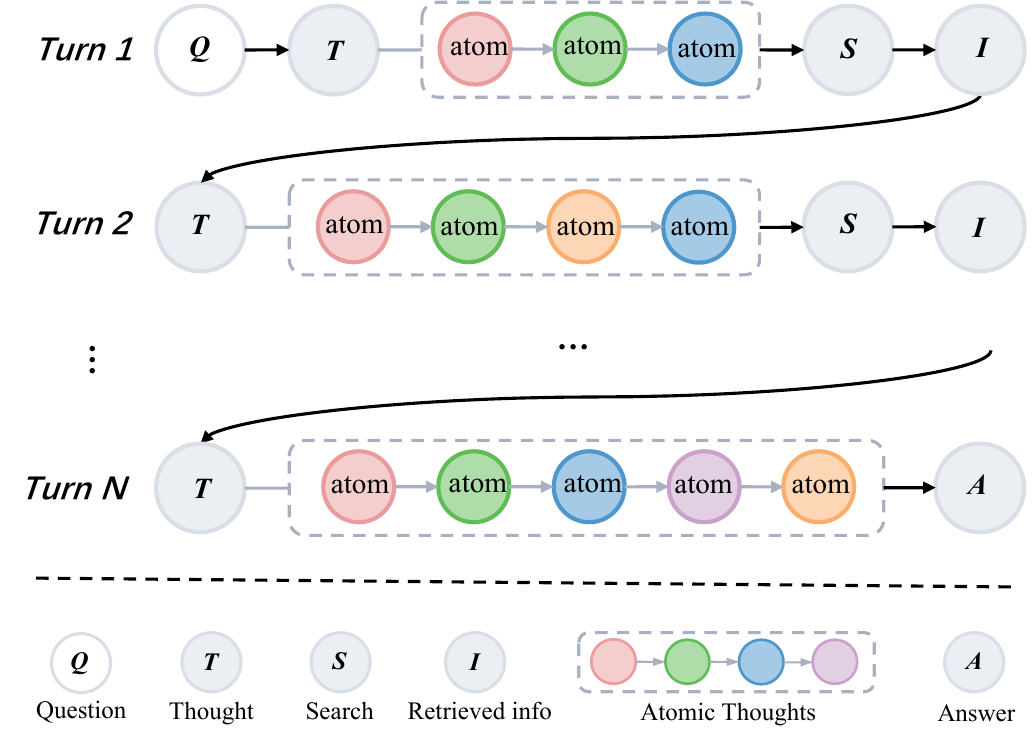}
  \vspace{-16pt}
  \caption{Atomic Thought paradigm automatically decomposes each  \(\texttt{<}\text{think}\texttt{>}\) into finer-grained functional units  \(\texttt{<}\text{atom-think}\texttt{>}\) during the rollout.}
  \label{fig:intro}
  \vspace{-10pt}
\end{wrapfigure}

Although large language models (LLMs) demonstrate impressive language understanding and logical reasoning abilities \cite{yang2025qwen3,guo2025deepseek,hurst2024gpt}, their capacity to solve complex problems ultimately hits a ceiling due to the static nature of their internal knowledge representation \cite{wang2024omnieval,jin2024long}. Retrieval-Augmented Generation (RAG) \cite{lewis2020retrieval} offers solution by equipping LLMs with external information sources, enhancing the relevance, accuracy, and timeliness of their responses \cite{gao2023retrieval,fan2024survey}. However, RAG's static workflows, making them ineffective at handling real-world questions that require sophisticated multi-hop reasoning and strategic search planning \cite{singh2025agentic}, as they often fail to construct correct search paths for complex problems \cite{yao2023tree}. To mitigate these limitations, a new search paradigm, termed \textbf{Agentic Deep Research} system, has been proposed, which enables autonomous reasoning, on-demand searching, and iterative information synthesis. Demonstrations from recent deep research systems by OpenAI \cite{openaideepresearch} and Google \cite{geminideepresearch} reveal several key advantages of this paradigm: 1) \textit{Comprehensive Understanding}: Effectively handles complex, multi-step queries that challenge traditional methods \cite{wei2022chain}; 2) \textit{Enhanced Synthesis}: Integrates diverse and even conflicting sources into coherent, informative outputs \cite{cheng2025survey}; 3) \textit{Reduced User Effort}: Automates tedious search processes, easing users’ cognitive and manual burden \cite{sami2024system}.

Early implementations of agentic deep research relied on prompt engineering \cite{song2024measuring,kim2024sure} and supervised fine-tuning (SFT) \cite{zhang2024scaling}. Yet, prompt-based methods rely heavily on LLMs’ instruction-following and long-context capabilities, whereas SFT tends to generalize poorly across domains \cite{chu2025sft}. More recently, post-training LLMs via reinforcement learning with outcome-based rewards (outcome-based RL) has yielded notable gains in reasoning performance \cite{guo2025deepseek, openaio1}. Building on this insight, recent advances \cite{webfilter, yang2025realfactbench, yang2025rama} (e.g. Search-R1 \cite{jin2025search} and DeepResearcher \cite{zheng2025deepresearcher}) treat the search tool as part of the environment and apply outcome-based RL to enable end-to-end optimization of the entire workflow, resulting in more performant and generalizable agentic deep research systems. Although outcome-based RL has shown promise, it remains insufficient in fully advancing agentic deep research, for the following reasons: 1) \textit{Gradients Conflicts}: In the outcome-based RL paradigm, an incorrect final answer results in the entire trajectory being penalized \cite{lightman2023let}, even when intermediate reasoning process or research strategies are effective. This coarse-grained reward design introduces potential gradient conflicts between intermediate reasoning steps and final answers, which hinders the model from discovering better reasoning capabilities and research strategies, thereby limiting its generalization ability. 2) \textit{Reward sparsity}: Outcome-based RL relies solely on the final answer to generate rewards \cite{du2024study}, resulting in each training sample providing only sparse feedback. This severely limits the efficiency of policy optimization, as it increases the reliance on larger training datasets and prolonged training schedules.

% 1) \textit{Suboptimal search calls}: Despite showing gains on leaderboard benchmarks, existing agentic deep research systems \cite{zheng2025deepresearcher, jin2025search, song2025r1} trained to convergence with outcome-based RL still exhibit frequent inefficient or failed search calls, indicating a lack of strategic competence in search tool utilization.  2) \textit{ Conflicting gradients}: In the outcome-based RL paradigm, an incorrect final answer results in the entire trajectory being penalized \cite{lightman2023let}, even when intermediate search strategies are effective. This leads to conflicting gradients that hinder the model from learning correct search strategies, thereby severely limiting generalization. 3) \textit{Reward sparsity}: Outcome-based RL relies solely on the final answer to generate rewards \cite{du2024study}, resulting in each training sample providing only sparse feedback. This, in turn, increases the dependence on larger training datasets and more training steps.

To address these challenges, we begin by introducing \textbf{Atomic Thought}, a novel LLM thinking paradigm that decomposes reasoning into fine-grained functional units, called Atomic Thoughts, guiding LLMs to engage in clearer and more in-depth reasoning, as illustrated in Figure~\ref{fig:intro}. 
% The Atomic Thoughts (units) are defined as minimal, functionally meaningful reasoning units obtained through a hierarchical abstraction of the reasoning process, as illustrated in Figure~\ref{fig:intro}
For example, reasoning operations like \(\texttt{<}\text{Reflection}\texttt{>}\) and \(\texttt{<}\text{Verification}\texttt{>}\) serve as Atomic Thoughts. Their interactions constitute the functional backbone of the reasoning process. To promote generalization, we avoid manual decomposition of Atomic Thoughts and instead encourage the model to autonomously induce them from reasoning processes. Building on this definition, we employ a Reasoning Reward Model (RRM) to score the generated Atomic thoughts and construct fine-grained \textbf{A}tomic \textbf{T}hought \textbf{R}eward (\textbf{ATR}). The ATR serves as an auxiliary signal to calibrate the outcome reward, thereby mitigating gradient conflicts during policy optimization. To aggregate the ATR and outcome reward, we design an curriculum-inspired strategy. During the early stages of training, the model is in a solution path exploration phase: while it may struggle to produce fully correct final answers, it can more easily develop partially correct reasoning traces. Relying solely on outcome rewards at this stage may induce severe gradient conflicts, thus requiring stronger calibration. As training advances, the alignment between reasoning and answers improves, reducing gradient conflicts and necessitating weaker calibration to avoid introducing excessive noise. Accordingly, we employ a linearly decaying weighting scheme, wherein the contribution of the ATR is gradually reduced as training proceeds. In addition, the hybrid reward incorporates process-level signals into the outcome-based reward, alleviating the problem of reward sparsity. Building on the above components, we propose \textbf{Atom-Searcher}, a novel RL framework for agentic deep research, aimed at advancing the performance frontier of agentic deep research models.

We conducted experiments on seven benchmarks covering both in-domain and out-of-domain tasks, demonstrating that Atom-Searcher achieves significant performance gains compared to the state-of-the-art (SOTA) baseline. Furthermore, we designed experiments to highlight the following advantages of Atom-Searcher: (1) Atom-Searcher effectively scales computation during test-time. (2) Atomic Thoughts provide supervision anchors for RRMs, effectively bridging deep research tasks and RRMs. (3) Atom-Searcher exhibits more interpretable, human-like reasoning patterns

In summary, our main contributions are as follows:
\begin{itemize}[leftmargin=2em]
    \item We first introduce Atomic Thought, a novel LLM thinking paradigm that decomposes reasoning into fine-grained functional units, effectively guiding LLMs to engage in clearer and more in-depth reasoning.
    \item Building on Atomic Thought, we design fine-grained Atomic Thought Reward and construct a curriculum-inspired aggregation strategy to integrate ATR with the outcome reward. This reward modeling alleviates gradient conflicts and reward sparsity during policy optimization.
    \item Building on Atomic Thought paradigm, ATR and the proposed reward aggregation strategy, we introduce Atom-Searcher, a novel RL framework for agentic deep research, aimed at advancing the performance frontier of agentic deep research.
    \item We demonstrated that Atom-Searcher achieves significant performance improvements over the SOTA baseline on seven benchmarks covering both in-domain and out-of-domain tasks. Additionally, we designed experiments to highlight a range of impressive advantages of Atom-Searcher.
\end{itemize}
\section{Atom-Searcher}
\begin{figure*}[!t]
\begin{center}
\centerline{\includegraphics[width=1.015\textwidth]{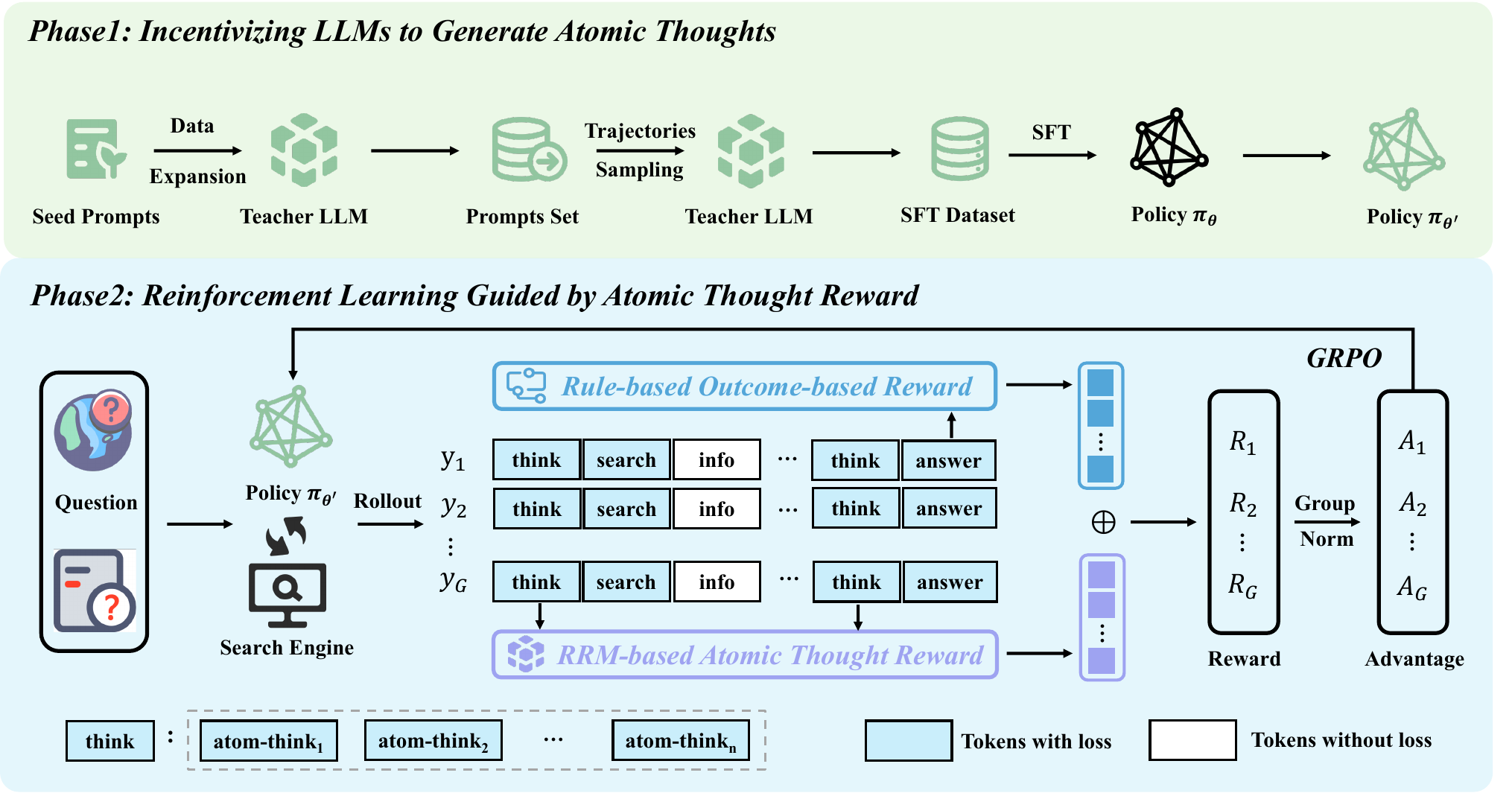}}
\vspace{-2mm}
\caption{
Overview of Atom-Searcher. Within the Atom-Searcher framework, we:
1) construct an atomic thought dataset and apply supervised fine-tuning (SFT) to the policy LLM—serving as the agentic deep research model—to incentivize its capability for generating atomic thoughts; 2) formulate fine-grained atomic thought rewards using a reasoning reward model, aligned with the atomic structure of the reasoning process, and integrate them with rule-based outcome rewards to optimize the SFT-initialized policy LLM via reinforcement learning.}
\label{fig:framework}
\end{center}
\vskip -0.3in
\end{figure*}
\label{method}
We propose a novel framework for enhancing agentic deep research models. As illustrated in Figure~\ref{fig:framework}, the framework consists of two phases. In phase1, we construct an atomic thought instruction dataset and perform SFT on the policy model to incentivize its ability to generate atomic thoughts. In phase2, we leverage a Reasoning Reward Model to derive fine-grained rewards based on the generated atomic thoughts, and integrate them with existing rule-based outcome rewards. The resulting hybrid reward is then used to further train the SFT-initialized policy LLM via RL.

\subsection{Preliminary}
\paragraph{\textbf{Atomic Thought}.} Understanding the fundamental units of thought is critical for simulating intelligence, optimizing decision-making, and extracting actionable knowledge in both cognitive science and computational reasoning \cite{anderson1997act, ho2022cognitive}. Drawing inspiration from philosophical conceptions of thought and the structured decomposition of actions in domains such as football, we propose a principled framework for defining the atomic thought within the reasoning processes of LLMs. A LLM atomic thought is the minimal, functionally coherent unit of reasoning, irreducible in form, yet integral to the model's reasoning trajectory. The interactions among atomic thoughts collectively form a functionally complete reasoning or behavior process. Take football as an example: when learning the kicking motion of a skilled player, we need to analyze the atomic units that compose this complex behavior, such as step adjustment, leg swing, and point of contact with the ball. Similarly, when shifting to LLMs, assessing the quality of their reasoning requires analyzing the atomic thoughts that compose their thought process. Therefore, in RL settings, designing fine-grained rewards at the atomic thought level can provide valuable intermediate supervision signals for guiding the reasoning trajectory.

In implementation, we encapsulate the LLM’s reasoning process within a \(\texttt{<}\text{atom-think}_i\texttt{>}\) tag and structure the atomic thoughts as subtags within it, as illustrated in Figure~\ref{fig:intro}. Importantly, the model is not constrained to follow manually defined atomic thoughts. Instead, we incentivize the model to autonomously generate atomic thoughts, enabling it to learn how to decompose reasoning into task-specific atomic thoughts across different scenarios.
% \[
%     \texttt{<}\text{think}\texttt{>} \ \texttt{<}\text{atom-think}_1\texttt{>}A_1\texttt{<}\text{/atom-think}_1\texttt{>} \dots \texttt{<}\text{atom-think}_n\texttt{>}A_n\texttt{<}\text{/atom-think}_n\texttt{>} \ \texttt{<}\text{/think}\texttt{>}
% \]
% where, \(\texttt{<}\text{atom-think}_i\texttt{>}\) denotes the atomic thought tag, such as \(\texttt{<}\text{Reflection}\texttt{>}\) or \(\texttt{<}\text{Verification}\texttt{>}\), and \(A_i\) represents the corresponding content of each atomic thought. 

\paragraph{\textbf{Atom-Searcher Trajectory}.}In an agentic deep research trajectory, the model iteratively performs reasoning and search invocations based on the user question and accumulated observations, as illustrated in Figure~\ref{fig:intro}) \textit{Reasoning}: Following the setup of DeepSeek-R1 \cite{guo2025deepseek}, we constrain Atom-Searcher to perform reasoning before taking any other action. Each segment of reasoning is encapsulated between the tags \(\texttt{<}\text{think}\texttt{>}\) and \(\texttt{<}\text{/think}\texttt{>}\). Notably, Atom-Searcher further decomposes the reasoning within the \(\texttt{<}\text{think}\texttt{>}\) tag into a sequence of atomic thoughts, each of which is encapsulated between the tags \(\texttt{<}\text{atom-think}\texttt{>}\) and \(\texttt{<}\text{/atom-think}\texttt{>}\) (e.g., \(\texttt{<}\text{Reflection}\texttt{>}\) and \(\texttt{<}\text{/Reflection}\texttt{>}\)). 2) \textit{Search}: After reasoning, Atom-Searcher may choose to invoke the web search tool by generating a JSON-formatted request with the tool name (web\_search) and the search queries as arguments. The request is encapsulated between the tags \(\texttt{<}\text{tool\_call}\texttt{>}\) and \(\texttt{<}\text{/tool\_call}\texttt{>}\). 3) \textit{Search Response}: When the system detects the tokens \(\texttt{<}\text{tool\_call}\texttt{>}\) and \(\texttt{<}\text{/tool\_call}\texttt{>}\), a search invocation is triggered. The retrieved results are then wrapped between the tags \(\texttt{<}\text{tool\_response}\texttt{>}\) and \(\texttt{<}\text{/tool\_response}\texttt{>}\) and appended to the current trajectory. 4) \textit{Answer}: Once Atom-Searcher determines that sufficient information has been gathered, it generates the final response enclosed between the tags \(\texttt{<}\text{answer}\texttt{>}\) and \(\texttt{<}\text{answer}\texttt{>}\). This serves as the final answer returned to the user.

\paragraph{\textbf{Problem Formulation}.}We model the process of completing the agentic deep research tasks as a finite-horizon Markov Decision Process (MDP), denoted by \((\mathcal{S},\mathcal{A},\mathcal{R},\mathcal{T})\). Given a user instruction \(I\), the agent is required to complete the corresponding task. The state \(s\in \mathcal{S}\) is defined as the retrieved content along with the history of previous actions. The action space \(\mathcal{A}\) includes three types of actions: 1) \(a^G\)(Generate Atomic Thought); 2) \(a^S\)(Invoke Search; and) 3) \(a^A\)(Answer). At the \(t\)-th step, conditioned on the state \(s_t\), the agent takes an action \(a_t\in \mathcal{A}\) following the LLM policy \(\pi_{\theta}\), which can be expressed as:
\begin{equation}
a_t = \pi_{\theta}(I,s_t)
\end{equation}

The agent then receives a reward \(r_t\) and the state is updated to \(s_{t+1}\). We formalize this process as follows.
\begin{equation}
    s_{t+1}=\mathcal{T}(s_t,a_t)
\end{equation}
\begin{equation}
\mathcal{T}(s_t,a_t)=
\begin{cases}
    \text{concat}(s_t; a_t, d_t) & \text{if } a_t=a_t^S \\
    \text{concat}(s_t;a_t) & \text{otherwise}
\end{cases}
\end{equation}
\begin{equation}
r_t=\mathcal{R}(s_t,a_t)
\end{equation}
where \(\mathcal{T}\) and \(\mathcal{R}\) denote the deterministic state transition function and deterministic reward function provided by the environment, respectively; \(\text{concat}(;)\) denotes the concatenation operation; \(d_t\) represents the retrieved external information; and \(r_t\) denotes the immediate reward at time step \(t\).
In the finite-horizon setting, the trajectory terminates either upon task completion or once the maximum number of interactions is reached.
Finally, based on the sampled trajectories, we optimize the policy \(\pi_{\theta}\) using the Group Relative Policy Optimization (GRPO) algorithm \cite{shao2024deepseekmath}. 
\subsection{Incentivizing LLMs to Generate Atomic Thoughts}
To enable LLMs to learn how to reasonably decompose their reasoning processes into atomic thoughts, we construct a high-quality atomic thought dataset \(D_{atom}\) consisting of 1,000 annotated examples and perform supervised fine-tuning to impart prior knowledge of atomic thought structures to the model. The details are as follows. The construction of \(D_{atom}\) involves two phases: 1) \textit{synthesizing atomic action prompts}: Firstly, we carefully design 10 distinct seed system prompt templates, each containing two atomic thought examples. Each example consists of 3 to 10 common atomic thoughts (e.g., \(\texttt{<}\text{plan}\texttt{>}\), \(\texttt{<}\text{reflection}\texttt{>}\), etc.). Secondly, we leverage a powerful teacher model (e.g., Qwen2.5-72B \cite{hui2024qwen2}) to generate approximately 1,000 system prompts based on the seed prompt templates, each containing a distinct combination of atomic thoughts. Finally, we combine each system prompt with different questions and callable search tools (e.g., web\_search) to obtain 1,000 prompts. 2) \textit{sampling high-quality reasoning trajectories}: Based on these 1,000 prompts, we use Qwen2.5-72B to sample complete reasoning trajectories. To ensure the quality of the generated trajectories, we employ a majority voting strategy during the sampling process. Data samples in \(D_{atom}\) follow the reasoning trajectory illustrated in Figure~\ref{fig:intro}. We perform SFT of \(\pi_{\theta}\) on \(D_{atom}\) to obtain \(\pi_{\theta^{\prime}}\), which is endowed with prior knowledge of atomic thoughts.
\subsection{Reward Modeling}
The introduction of atomic thoughts offers a promising perspective for designing fine-grained reward signals to guide agentic deep research models toward developing more intelligent and efficient research strategies. We first construct ATR using a reasoning reward model, and then integrate them with the outcome-level reward through a training-dynamics-aware, linearly decaying aggregation strategy.
\paragraph{\textbf{Constructing fine-grained atomic thought reward}.}
With the rapid progress of foundation model capabilities and the rise of test-time scaling techniques \cite{snell2024scaling}, Reasoning Reward Models (RRMs) \cite{liu2025inference}, which leverage large reasoning models (e.g., DeepSeek-R1 \cite{guo2025deepseek}) to generate rewards, have become a promising solution. RRMs are particularly effective in settings that require fine-grained supervision, adaptive reasoning, and open-ended tasks without ground truth, making them well aligned with the characteristics of atomic thoughts. Therefore, we use the RRM to score the atomic thoughts generated by the policy model, resulting in the ATR. This process can be formulated as follows::
\begin{equation}
    r_{atom}^1,r_{atom}^2,...,r_{atom}^n=RRM(I_{score}, y)
\end{equation}
\begin{equation}
    R_{atom}=f(r_{atom}^1,r_{atom}^2,...,r_{atom}^n)
\end{equation}
where, \(I_{score}\) denotes the scoring prompt, as illustrated in Figure~\ref{fig:RRM Atom Prompt}; \(y\) refers to the generated trajectory; \(r_{atom}^i\) represents the score of the \(i\)-th atomic thought, and \(f(\cdot)\) denotes the aggregation function that combines individual atomic scores. The choice of \(f(\cdot)\) is not fixed—it can be a simple average or a more sophisticated weighting strategy. \(R_{atom}\) denotes the ATR of trajectory \(y\).
\paragraph{\textbf{A Dynamic, Curriculum-Based Approach to Reward Aggregation}.}A key limitation of outcome-based reward is their coarse credit assignment: it attribute the correctness of intermediate reasoning solely to the final answer, often rewarding or penalizing steps regardless of their actual contribution. This misalignment introduces gradient conflicts during optimization. To address this, we aggregate ATR with the outcome reward, using ATR as an auxiliary signal to calibrate the final reward, thereby mitigating gradient conflicts and improving test-time performance. However, using a static weighting coefficient for reward aggregation fails to align with training dynamics. Specifically, early in training, the model—still limited in its deep research capability—struggles to generate fully correct answers but is more likely to explore useful atomic thoughts that contribute toward a correct solution. If training relies solely on outcome-based rewards at this stage, these beneficial atomic thoughts may be unjustly penalized due to the incorrect final answer; conversely, harmful atomic thoughts may also be mistakenly reinforced, resulting in severe gradient conflict and necessitating strong calibration from ATR. As training progresses and the model’s deep research ability improves, its reasoning trajectories become increasingly aligned with correct answers. Consequently, gradient conflicts diminish, and excessive calibration from ATR may introduce unnecessary noise, potentially harming final accuracy. To accommodate this, we adopt a training-dynamics-aware weighting scheme that linearly reduces the contribution of ATR as training progresses, formulated mathematically as follows:
\begin{equation}
    \alpha=0.5 \times (1-\frac{T}{T_{MAX}})
\end{equation}
\begin{equation}
    R = 
    \begin{cases}
        \alpha R_{atom}+(1-\alpha )R_{f1} & \text{if format is correct}  \\
        -1 & \text{if format is incorrect}
    \end{cases}
\end{equation}
\begin{equation}
    R_{f1}=\frac{2\times IN}{PN+RN}
\end{equation}
where, \(T\) denotes the current training step, and \(T_{MAX}\) denotes the maximum number of training steps. \(R\) denotes the final reward used for RL training, and \(R_{f1}\) represents the outcome-based reward computed from the F1 score. The coefficient \(\alpha \in [0,1]\) is a hyperparameter that balances the influence of ATR and the outcome reward during training. \(PN\) denotes the word count of the predicted answer, \(RN\) denotes the word count of the reference answer, and \(IN\) denotes the word count of their intersection.
\subsection{RL Training Framework}
\paragraph{\textbf{Policy Optimization}.} 
In this work, we adopt the GRPO algorithm \cite{shao2024deepseekmath} to optimize the SFT policy \(\pi_{\theta^{\prime}}\) using the hybrid reward \(R\) that aggregates final answer correctness and reasoning quality. GRPO improves the current policy \(\pi_{\theta^{\prime}}\) by leveraging a reference policy \(\pi_{\theta^{\prime}_{\text{ref}}}\) and a set of rollouts generated by a previous policy \(\pi_{\theta^{\prime}_{\text{old}}}\). The objective is extended and formulated as follows:
\begin{equation}
    r_1,r_2,...,r_G=R(y_1,y_2,...,y_G)
\end{equation}
\begin{equation}
    A_i=\frac{r_i-mean(r_1,r_2,...,r_G)}{std(r_1,r_2,...,r_G)}    
\end{equation}
\begin{equation}
\label{GRPO}
\begin{aligned}
\mathcal{J}_{GRPO}(\theta^{\prime}) =\ & \mathbb{E}_{x \sim \mathcal{D},\, \{y_i\}_{i=1}^{G} \sim \pi_{\theta^{\prime}_{\text{old}}}(\cdot|x)} \Bigg[ \frac{1}{G} \sum_{i=1}^{G} \min \Bigg( \frac{\pi_{\theta^{\prime}}(y_i|x)}{\pi_{\theta^{\prime}_{\text{old}}}(y_i|x)} A_i, \\[4pt]
& \text{clip} \left( \frac{\pi_{\theta^{\prime}}(y_i|x)}{\pi_{\theta^{\prime}_{\text{old}}}(y_i|x)},\ 1 - \epsilon,\ 1 + \epsilon \right) A_i \Bigg) - \beta\, \mathbb{D}_{\mathrm{KL}}\left(\pi_{\theta^{\prime}} \,\|\, \pi_{\theta^{\prime}_{\mathrm{ref}}}\right) \Bigg]
\end{aligned}
\end{equation}
where \(x\) denotes an input sampled from the experience distribution \(D\), \(y_i\) represents a trajectory generated by \(\pi_{\theta_{\text{old}}^{\prime}}\), \(G\) is the number of trajectories sampled per training example, \(r_i\) is the reward of \(y_i\), \(A_i\) is the advantage of \(y_i\), \(\mathbb{D}_{\mathrm{KL}}\) denotes the unbiased estimate of KL divergence \cite{shao2024deepseekmath}, and \(\beta\) is a tunable hyperparameter. In addition, to mitigate entropy collapse during policy optimization, we adopt a sliding-window-based entropy regulation mechanism, as detailed in Appendix \ref{A.1}.
\paragraph{\textbf{Loss Masking}.}In the original GRPO framework, loss is computed over all tokens in the trajectory. However, in Atom-Searcher, trajectories include retrieval results that are externally fetched by the environment rather than generated by the policy itself. To prevent biasing the policy update toward non-trainable, static content, we apply loss masking to exclude these retrieved segments from the optimization objective. Specifically, in the computation of Equation \ref{GRPO}, only tokens corresponding to the model's reasoning (i.e., text-based thinking) and search queries are included, while tokens originating from retrieval results are masked out.
\section{Experiments}
\label{Exp}
\subsection{Implementation Details}
We use Qwen2.5-7B-Instruct \cite{qwen2025qwen25technicalreport} as the backbone models. The training is conducted using the verl framework \cite{sheng2024hybridflow}. At each training step, we sample 32 prompts and generate 16 rollouts per prompt. Each rollout consists of up to 10 tool calls, followed by a final answer step. The training is performed with a mini-batch size of 512, meaning that one rollout stage corresponds to a single backpropagation step. By default, we use Qwen3-30B-A3B \cite{yang2025qwen3} as the reasoning reward model in Atom-Searcher.
\subsection{Benchmarks}
To comprehensively assess model performance in both in-domain (ID) and out-of-domain (OOD) scenarios, we construct a diverse evaluation benchmark spanning a wide range of open-domain QA tasks. For ID evaluation, we include the development sets of NQ \cite{kwiatkowski2019natural}, TQ \cite{joshi2017triviaqa}, HotpotQA \cite{yang2018hotpotqa}, and 2Wiki \cite{ho2020constructing}. To evaluate OOD generalization, we incorporate three datasets that differ substantially in question format and information distribution: MuSiQue \cite{trivedi2022musique}, Bamboogle \cite{press2022measuring}, and PopQA \cite{mallen2022not}. These datasets are chosen to challenge the model’s ability to generalize beyond its training distribution.

To ensure fair comparison and balanced evaluation, we randomly sample 512 examples from the development sets of NQ, TQ, HotpotQA, 2Wiki, MuSiQue, and PopQA, along with all 125 examples from the Bamboogle development set. This evaluation setup enables a rigorous assessment of model robustness across diverse topics and reasoning demands.
\subsection{Baselines}
To evaluate the effectiveness of Atom-Searcher, we compare it against the following baseline methods:
\begin{itemize}[leftmargin=2em]
    \item \textbf{CoT}: This baseline performs Chain-of-Thought (CoT) reasoning to generate answers without access to any external reference context.
    \item \textbf{Cot+RAG}: This baseline integrates CoT reasoning with retrieved reference context to guide the answer generation process.
    \item \textbf{Search-o1}: This baseline performs multi-step reasoning by generating search queries or intermediate answers. For each query, the model receives only a snippet retrieved by a retriever, rather than the full document content.
    \item \textbf{Search-o1-Web}: Unlike Search-o1, this setting allows the model to interact with the open web by issuing real-time queries through APIs and browsing webpages via URLs. This capability supports more dynamic and comprehensive information acquisition, laying the groundwork for deep research.
    \item \textbf{Search-r1}: This is a reinforcement learning approach for question answering that utilizes a retriever to search Wikipedia during both training and inference. It includes two variants—Search-r1-base and Search-r1-instruct—which are initialized from either the base model or the instruct-tuned model, respectively. 
    \item \textbf{R1-Seaecher}: This is a two-stage, outcome-driven RL baseline that equips LLMs with autonomous search: the model learns to invoke external search tools and incorporate retrieved evidence on the fly to improve reasoning.
    \item \textbf{DeepResearcher}: This is an end-to-end trained LLM agent for deep research tasks, leveraging reinforcement learning in real-world web environments. It interacts with the open web via real-time search and browsing, enabling dynamic information acquisition. 
\end{itemize}

\begin{table*}[h]
\centering
\caption{
Performance comparison of Atom-Searcher and baselines on in-domain and out-of-domain benchmarks, evaluated by F1 score; the best and second-best results are marked in \textbf{bold} and \underline{underlined}, respectively.
}
\label{tab: main}

\resizebox{1\textwidth}{!}{
% \begin{tabular}{
%     ll
%     S[table-format=2.1] S[table-format=2.1]
%     S[table-format=2.1] S[table-format=2.1]
%     S[table-format=2.1] S[table-format=2.1]
%     S[table-format=2.1] S[table-format=2.1]
% }
%\large
\setlength{\tabcolsep}{4pt} % 减小列间距，避免过宽
\renewcommand{\arraystretch}{1.2}

% 计算：总宽度约 \textwidth，共 8 个数据列，每列宽度 ≈ 0.11\textwidth
\newcolumntype{C}{>{\centering\arraybackslash}w{c}{0.11\textwidth}}
\fontsize{11pt}{12pt}\selectfont
\begin{tabular}{l l C C C C C C C}
\toprule
\multirow{2}{*}{\centering\textbf{Type}} & \multirow{2}{*}{\centering\textbf{Method}}
& \multicolumn{4}{c}{\textbf{In-domain}} 
& \multicolumn{3}{c}{\textbf{Out-of-domain}}  \\ 
\cmidrule(lr){3-6}
\cmidrule(lr){7-9}

& & {\textbf{NQ}} & {\textbf{TQ}} 
  & {\textbf{HotpotQA}} & {\textbf{2Wiki}} 
  & {\textbf{Musique}} & {\textbf{Bamboogle}} 
  & {\textbf{PopQA}} \\ 
\midrule
\multirow{4}{1.5cm}{\textbf{\textit{Prompt Based}}} 
& CoT          & 19.8 & 45.6 & 24.4 & 26.4 & 8.5 & 22.1 & 17.0  \\
& CoT+RAG   & 42.0 & 68.9 & 37.1 & 24.4 & 10.0 & 25.4 & 46.9  \\
& Search-o1  & 34.5 & 52.6 & 31.6 & 28.6 & 16.8 & 35.8 & 36.9  \\
& Search-o1-Web  & 32.4 & 58.9 & 33.0 & 30.9 & 14.7 & 46.6 & 38.3  \\
\midrule
\multirow{5}{1.5cm}{\textbf{\textit{Training Based}}} 
& Search-r1-base          & \textbf{45.4} & 71.9 & \underline{55.9} & 44.6 & 26.7 & 56.5 & 43.2 \\
& Search-r1-Instruct     & 33.1 & 44.7 & 45.7 & 43.4 & 26.5 & 45.0 & 43.0 \\
& R1-Searcher & 35.4 & 73.1 & 44.8 & 59.4 & 22.8 & 64.8 & 42.7 \\
& DeepResearcher & 39.6 & \underline{78.4} & 52.8 & \underline{59.7} & \underline{27.1} & \textbf{71.0} & \underline{48.5} \\
& \textbf{Atom-Searcher} & \underline{44.0} & \textbf{81.8} & \textbf{57.3} & \textbf{66.9} & \textbf{27.6} & \underline{70.7} & \textbf{50.3} \\
% \midrule
% \multirow{3}{*}{Web Search} 
% & R1-Searcher        & 35.4 & 52.3 & 73.1 & 79.1 & 44.8 & 53.1 & 59.4 & 65.8 \\
% & DeepResearcher     & 39.6 & 61.9 & \textbf{78.4} & 85.0 & 52.8 & 64.3 & 59.7 & 66.6 \\
% & \textbf{WebFilter (Ours)} 
%                      & \textbf{40.1} & \textbf{63.1} 
%                      & 77.5 & \textbf{85.4} 
%                      & \textbf{55.1} & \textbf{65.2} 
%                      & \textbf{60.1} & \textbf{67.5} \\
\bottomrule
\end{tabular}
}
\end{table*}

\begin{table*}[h]
\centering
\caption{
Ablation study of Atom-Searcher on seven QA benchmarks. We analyze the contribution of each component (RRM and Atom Thought). The \textbf{bold} indicates the best performance, and \underline{underline} indicates the second-best performance.
}
\label{tab:ablation study}

\resizebox{1\textwidth}{!}{
% \begin{tabular}{
%     ll
%     S[table-format=2.1] S[table-format=2.1]
%     S[table-format=2.1] S[table-format=2.1]
%     S[table-format=2.1] S[table-format=2.1]
%     S[table-format=2.1] S[table-format=2.1]
% }
%\large
\setlength{\tabcolsep}{4pt} % 减小列间距，避免过宽
\renewcommand{\arraystretch}{1.2}

% 计算：总宽度约 \textwidth，共 8 个数据列，每列宽度 ≈ 0.11\textwidth
\newcolumntype{C}{>{\centering\arraybackslash}w{c}{0.11\textwidth}}
\fontsize{10pt}{12pt}\selectfont
\begin{tabular}{l C C C C C C C}
\toprule
\multirow{2}{*}{\centering\textbf{Method}}
& \multicolumn{4}{c}{\textbf{In-domain}} 
& \multicolumn{3}{c}{\textbf{Out-of-domain}}  \\ 
\cmidrule(lr){2-5}
\cmidrule(lr){6-8}

& {\textbf{NQ}} & {\textbf{TQ}} 
  & {\textbf{HotpotQA}} & {\textbf{2Wiki}} 
  & {\textbf{Musique}} & {\textbf{Bamboogle}} 
  & {\textbf{PopQA}} \\ 

\midrule
Base & 39.6 & \underline{78.4} & 52.8 & 59.7 & \underline{27.1} & \textbf{71.0} & 48.5 \\
+ RRM & \underline{40.1} & 78.2 & \underline{53.5} & \underline{60.0} & 25.7 & 70.5 & \underline{48.8} \\
Atom-Searcher & \textbf{44.0} & \textbf{81.8} & \textbf{57.3} & \textbf{66.9} & \textbf{27.6} & \underline{70.7} & \textbf{50.3} \\
% \midrule
% \multirow{3}{*}{Web Search} 
% & R1-Searcher        & 35.4 & 52.3 & 73.1 & 79.1 & 44.8 & 53.1 & 59.4 & 65.8 \\
% & DeepResearcher     & 39.6 & 61.9 & \textbf{78.4} & 85.0 & 52.8 & 64.3 & 59.7 & 66.6 \\
% & \textbf{WebFilter (Ours)} 
%                      & \textbf{40.1} & \textbf{63.1} 
%                      & 77.5 & \textbf{85.4} 
%                      & \textbf{55.1} & \textbf{65.2} 
%                      & \textbf{60.1} & \textbf{67.5} \\
\bottomrule
\end{tabular}
}
\end{table*}

\subsection{Main Result}
Our main result, presented in Table \ref{tab: main}, show that Atom-Searcher achieves significant performance gains over both prompt-based and training-based baselines on in-domain and out-of-domain benchmarks. 
\subsubsection{Atom-Searcher outperforms baselines on in-domain benchmarks}
In the in-domain results, Atom-Searcher achieved the best performance on the TQ, HotpotQA and 2Wiki benchmarks, showing significant improvements over the second-best results, with increases of 4.3\%, 2.5\% and 12.1\%, respectively. On average, Atom-Searcher outperformed the SOTA baseline (DeepResearcher) by 8.5\% across the four in-domain benchmarks. Notably, while Search-r1-base achieved optimal performance on NQ, it was trained and evaluated using a local RAG system with direct access to the relevant Wikipedia corpus. In contrast, Atom-Searcher navigates the entire Internet to find relevant information, presenting a more realistic and challenging scenario, despite both models ultimately sourcing answers from Wikipedia.
\subsubsection{Atom-Searcher demonstrates optimal out-of-domain generalization}
In the out-of-domain results, Atom-Searcher achieved the best performance on the Musique and PopQA benchmarks, improving over the second-best performance by 1.8\% and 3.7\%, respectively. On Bamboogle, it achieved second-best performance, but was only 0.4\% lower than the optimal result. On average, Atom-Searcher outperformed the SOTA baseline (DeepResearcher) by 2.5\% across the three out-of-domain benchmarks. This demonstrates that Atom-Searcher effectively generalizes the skills learned during RL to unseen scenarios.
\subsection{Atom-Searcher Effectively Scales Computation at Test Time}
% \begin{table}[h]  % 改为 table（单栏），原来是 table*
% \centering
% \caption{
% Ablation study of Atom-Searcher on seven QA benchmarks. We analyze the contribution of each component (RRM and Atom Thought). The \textbf{bold} indicates the best performance, and \underline{underline} indicates the second-best performance.
% }
% \label{tab:ablation_study}

% \resizebox{1\textwidth}{!}{%
% \setlength{\tabcolsep}{8pt}
% \renewcommand{\arraystretch}{1.3}

% % 定义等宽列，m{} 类型已支持垂直居中
% \newcolumntype{C}{>{\centering\arraybackslash}m{0.22\textwidth}} % 调整宽度以适应单栏

% \fontsize{11pt}{12pt}\selectfont
% \begin{tabular}{C C C C}
% \toprule
% \multirow{2}{*}{\textbf{Method}}  % 纵向居中
% & \makecell{\textit{avg.\#} \\ \textbf{response} \\ \textbf{tokens}}
% & \makecell{\textit{avg.\#} \\ \textbf{think} \\ \textbf{tokens}}
% & \makecell{\textit{avg.\#} \\ \textbf{tool} \\ \textbf{calls}} \\
% \midrule
% DeepResearcher & 39.6 & \underline{78.4} & 52.8 \\
% Atom-Searcher  & \textbf{43.8} & \textbf{81.8} & \textbf{55.7} \\
% \bottomrule
% \end{tabular}%
% }
% \end{table}
\newcolumntype{C}{>{\centering\arraybackslash}m{0.22\textwidth}}
\begin{wraptable}{r}{0.48\textwidth}
  \vspace{-18pt} % 调整顶部间距
  \centering
  % \caption{The token generation statistics for Atom-Searcher and DeepResearcher during testing. "\textit{avg.\#} \textbf{response} \textbf{tokens}" refers to the average number of tokens in a complete response, "\textit{avg.\#} \textbf{think} \textbf{tokens}" represents the average number of tokens in a single think process within the response, and "\textit{avg.\#} \textbf{tool} \textbf{calls}" indicates the average number of tool calls in the response.}
  \caption{Test-time token generation statistics for Atom-Searcher vs DeepResearcher.}
  \vspace{-10pt}
  \label{tab:test-time scaling}
  \resizebox{\linewidth}{!}{%
  \setlength{\tabcolsep}{8pt}
  \renewcommand{\arraystretch}{1.3}
  \large
  \fontsize{16pt}{14pt}\selectfont
  \begin{tabular}{C C C C}
  \toprule
  \multirow{2}{*}{\textbf{Method}} 
  & \makecell{\textit{avg.\#} \\ \textbf{response} \\ \textbf{tokens}}
  & \makecell{\textit{avg.\#} \\ \textbf{think} \\ \textbf{tokens}}
  & \makecell{\textit{avg.\#} \\ \textbf{tool} \\ \textbf{calls}} \\
  \midrule
  \textbf{DeepResearcher} & 176 & 55 & 2.13 \\
  \mbox{\textbf{Atom-Searcher}}  & 565 & 143 & 2.65 \\
  \bottomrule
  \end{tabular}%

  }
 % 调整底部间距
   \vspace{-10pt}
\end{wraptable}
To analyze whether Atom-Searcher can effectively scale computation at test time, we compared the average number of tokens generated during the testing phase between Atom-Searcher and the SOTA baseline DeepResearcher. As shown in Table \ref{tab:test-time scaling}, Atom-Searcher generates 3.2 times more tokens in the average response length (\textit{avg.\#} \textbf{response} \textbf{tokens}) compared to DeepResearcher. In terms of the average length of a single think process within the response (\textit{avg.\#} \textbf{think} \textbf{tokens}), Atom-Searcher generates 2.6 times more tokens. Additionally, Atom-Searcher performs 1.24 times more tool calls per response (\textit{avg.\#} \textbf{tool} \textbf{calls}) than DeepResearcher. This demonstrates that the Atom-Searcher architecture effectively achieves Test-Time Scaling without the introduction of additional incentives for generating more tokens, highlighting its stronger exploration and discovery capabilities when handling complex and challenging deep research tasks.

\subsection{Ablation Study}
We conduct an ablation study to evaluate the impact of the Atomic Thought and the fine-grained rewards generated by RRM on Atom-Searcher. To assess their contributions, we compare \textbf{Atom-Searcher} with two alternative frameworks: (1) \textbf{Base} refers to the DeepResearcher \cite{zheng2025deepresearcher} setting, indicating Atom-Searcher w/o Atomic Thought \& fine-grained rewards generated by RRM. (2) \textbf{+RRM} refers to the incorporation of fine-grained rewards generated by RRM (with the same implementation details as Atom-Searcher) on top of the Base setting, indicating Atom-Searcher w/o Atomic Thought. As shown in Table \ref{tab:ablation study}, the results across seven benchmarks, including both in-domain and out-of-domain, indicate that \textbf{+RRM} does not yield a significant performance improvement over \textbf{Base}. This suggests that directly using RRM for fine-grained supervision provides minimal benefits. However, \textbf{Atom-Searcher} significantly outperforms \textbf{+RRM}, achieving an average performance improvement of 6.1\% across four in-domain benchmarks and 2.5\% across three out-of-domain benchmarks, demonstrating the contribution of Atomic Thought. The above results raise an interesting question: \textbf{why does direct supervision using RRM have minimal effect on the reasoning process, while its effectiveness significantly improves after decomposing the reasoning process into Atom Thoughts}? We speculate that this is because \textbf{Atom Thoughts provide supervision anchors for RRM, helping it focus on the effective functional modules in the reasoning process, thereby generating meaningful fine-grained reward signals} (ATR in Atom-Searcher).

\subsection{Case Study}
\begin{figure*}[!t]
\begin{center}
\centerline{\includegraphics[width=1\textwidth]{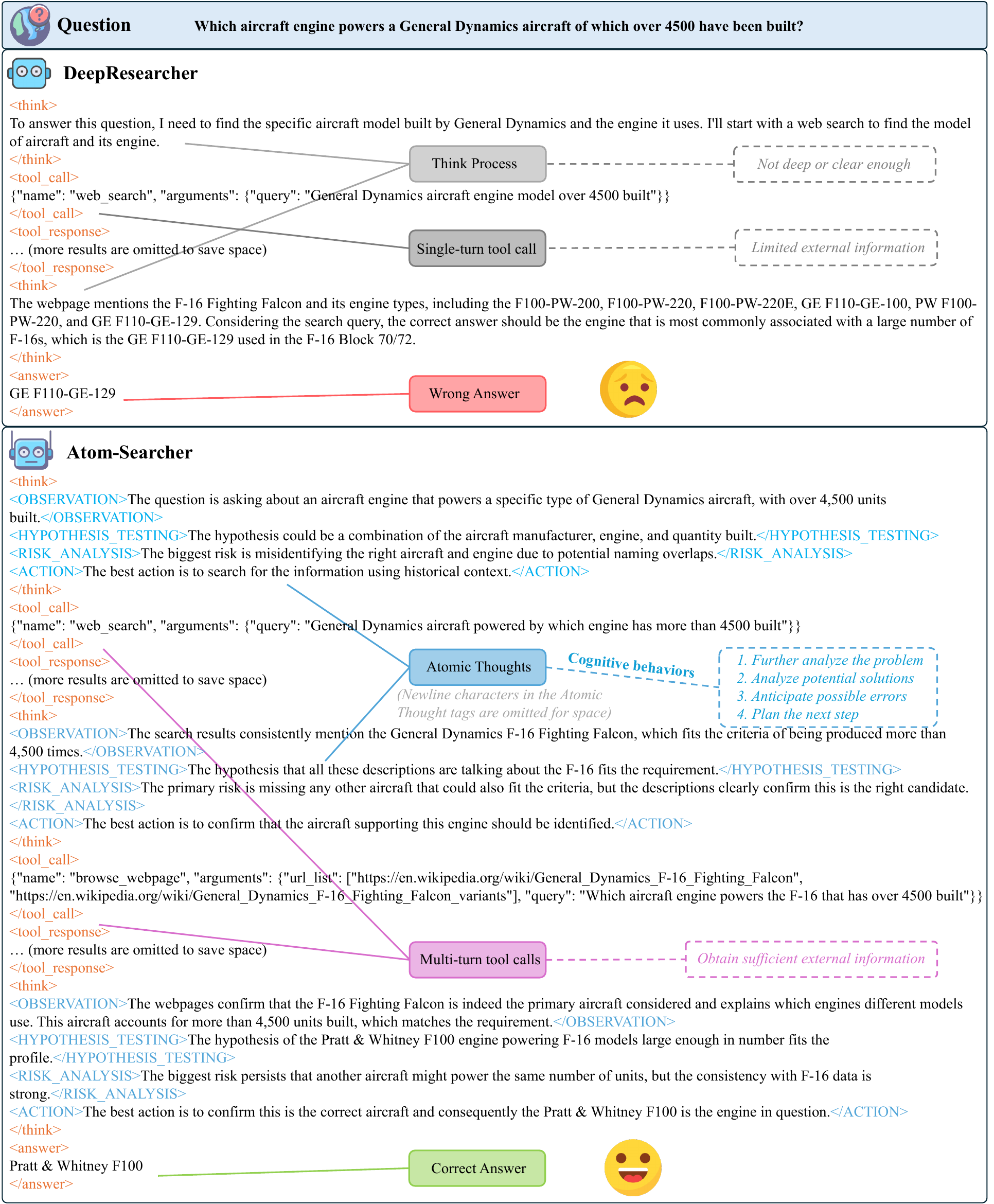}}
\vspace{-2mm}
\caption{
The case study demonstrates a comparison of the reasoning behavior between Atom-Searcher (below) and the SOTA baseline DeepResearcher (above).}
\label{fig:case_study}
\end{center}
\vskip -0.3in
\end{figure*}

\begin{figure*}[!t]
\begin{center}
\centerline{\includegraphics[width=1\textwidth]{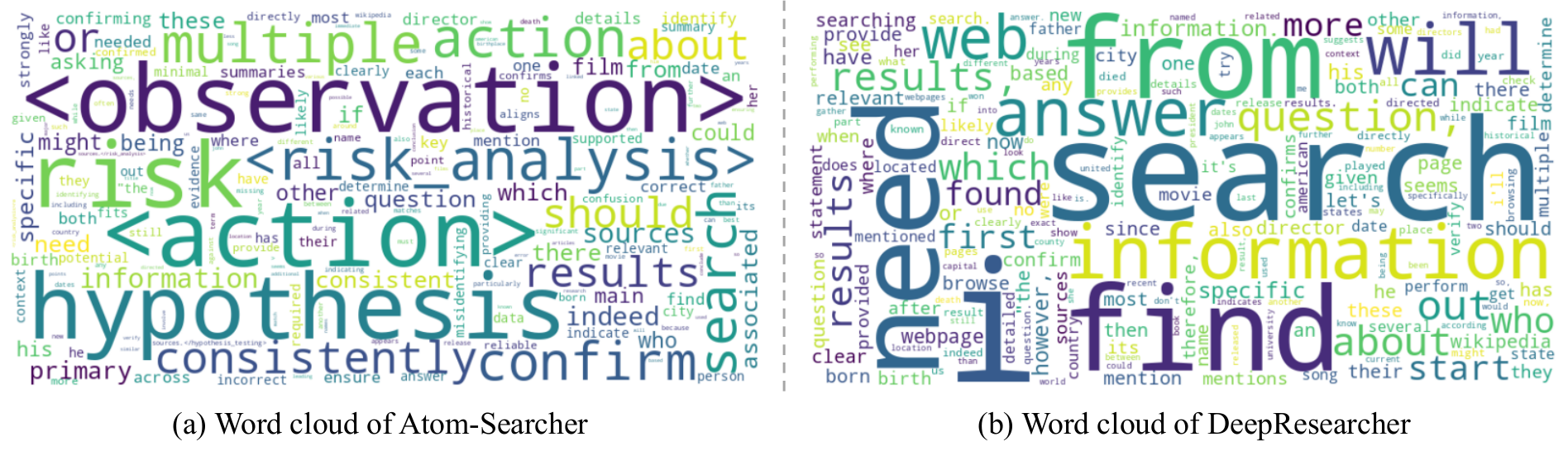}}
\vspace{-2mm}
\caption{
Word cloud: Token frequency statistics of the responses during the testing phase for Atom-Searcher (a) and DeepResearcher (b).}
\label{fig:word_cloud}
\end{center}
\vskip -0.3in
\end{figure*}

Figure \ref{fig:case_study} analyzes the behavioral differences between Atom-Searcher and the SOTA baseline DeepResearcher in completing a deep research task. It demonstrates the following advantages of Atom-Searcher: (1) Atom-Searcher employs Atomic Thoughts in its reasoning, which leads to more human-like cognitive behaviors, such as problem analysis, solution hypotheses, error prediction, and next-step planning, making its reasoning process deeper and clearer. (2) Atom-Searcher triggers more search calls, allowing it to obtain richer external information to ensure the correctness of the answer. These advantages indicate that Atom-Searcher has great potential in more complex deep research tasks.

Additionally, we analyzed the token frequency statistics for Atom-Searcher and DeepResearcher during the testing phase. The word cloud, shown in Figure \ref{fig:word_cloud}, illustrates the most frequently occurring tokens. The top-5 most frequent tokens in Atom-Searcher are \texttt{<observation>}, \texttt{<action>}, \texttt{hypothesis}, \texttt{risk}, and \texttt{<risk\_analysis>}, whereas the top-5 most frequent tokens in DeepResearcher are \texttt{I}, \texttt{search}, \texttt{need}, \texttt{find}, and \texttt{from}. This disparity suggests that, compared to DeepResearcher, Atom-Searcher better aligns with human-like efficient cognitive patterns when performing deep research tasks, with a stronger focus on in-depth problem analysis, hypothesis evaluation, risk assessment, and strategic planning.
\section{Related Work}
\subsection{Prompt and SFT-based Agentic Deep Research}
Early prompt-based paradigms rely on human-authored workflows to specify the interaction between LLMs and external knowledge sources. \cite{wang2024searching}. For example, OpenResearcher \cite{zheng2024openresearcher}, AirRAG \cite{feng2025airrag}, IterDRAG \cite{yue2024inference}, Plan*RAG \cite{verma2025plan}, Search-o1 \cite{li2025search}, and Open Deep Search \cite{alzubi2025open} have advanced search capabilities via carefully designed workflows. However, their reliance on human-engineered prompts and interaction patterns imposes rigid behavior constraints, limiting adaptability. These limitations motivate a shift toward SFT-based approaches that support more flexible and adaptive search strategies \cite{yu2024auto, wang2024corag}. For example, CoRAG \cite{wang2024corag} employs Monte Carlo Tree Search (MCTS) to dynamically select document blocks under budget constraints. However, it suffers from high computational overhead and limited generalization to unseen scenarios due to its reliance on supervised signals.
\subsection{RL-based Agentic Deep Research}
As LLMs have achieved remarkable breakthroughs in reasoning through outcome-based RL \cite{guo2025deepseek, team2025kimi}, this paradigm is emerging as a promising direction for enhancing agentic deep research via end-to-end optimization, attracting growing interest and active exploration from the research community. Recent works, such as ReSearch \cite{chen2025learning}, Search-R1 \cite{jin2025search}, R1-Searcher \cite{song2025r1}, DeepResearcher \cite{zheng2025deepresearcher}, WebRL \cite{qi2024webrl}, WebThinker \cite{li2025webthinker}, ZeroSearch \cite{sun2025zerosearch} and WebAgent-RL \cite{wei2025webagent} have extended outcome-supervised reinforcement learning to the agentic deep research setting, enabling LLMs to autonomously leverage search engines for complex reasoning tasks. Although enhancing agentic deep research with outcome-supervised reinforcement learning has led to performance gains, the coarse-grained reward signals provide limited guidance for learning efficient and intelligent search strategies, often resulting in suboptimal search calls. To overcome this, we propose an atomic thought-aware fine-grained reward to guide the model toward more efficient and intelligent search behaviors, while mitigating the training inefficiency caused by reward sparsity. Building on this, we further introduce a novel agentic deep research framework, Atom-Searcher, that integrates this reward formulation into a reinforcement learning paradigm.
\section{Conclusion}
\label{conclusion}
In this work, we first introduce Atomic Thought, a novel LLM thinking paradigm designed to guide LLMs in clearer and more in-depth reasoning. We then supervise Atomic Thoughts using a Reasoning Reward Model to generate fine-grained Atomic Thought Reward and aggregate them with outcome reward through a training-dynamics-aware strategy. Based on this, we propose Atom-Searcher, a novel RL framework for agentic deep research, which advances the performance frontier of agentic deep research models by addressing the conflicting gradients and reward sparsity issues present in existing outcome-based deep research frameworks. Experimental results demonstrate the outstanding performance of Atom-Searcher and a range of impressive advantages.

\bibliographystyle{assets/plainnat}
\bibliography{ref/Top,ref/reference}
\appendix
\section{Training Details}
\subsection{Sliding-Window-based Entropy Regulation Mechanism}
\label{A.1}
A major obstacle in scaling reinforcement learning for LLMs is the occurrence of entropy collapse \cite{yu2025dapo}, characterized by a rapid sharp drop in policy entropy at the early training stage, which results in an overconfident policy and severely impairs exploration. To  mitigate policy entropy collapse, we introduce a \textbf{S}liding-\textbf{W}indow-based dynamic \textbf{E}ntropy \textbf{R}egulation \textbf{M}echanism (\textbf{SWERM}) applied at the granularity of training steps. Before introducing SWERM, we first define policy entropy \(\mathcal{H}\) as the average token-level entropy of the policy model \(\pi_{\theta^{\prime}}\) over the current batch \(\mathcal{B}\), which can be formulated as follows:
\begin{equation}
    \mathcal{H}(\pi_{\theta^{\prime}},\mathcal{B})=-\mathbb{E}_{\mathcal{B},\pi_{\theta^{\prime}}}[\log\pi_{\theta^{\prime}}(y_t|\boldsymbol{y}_{<t})]=-\frac{1}{|\mathcal{B}|}\sum_{x\in \mathcal{B}}\frac{1}{|\boldsymbol{y}|}\sum_{t=1}^{|\boldsymbol{y}|}\mathbb{E}_{y_t\sim \pi_{\theta^{\prime}}}[\log\pi_{\theta^{\prime}}(y_t|\boldsymbol{y}_{<t},x)]
\end{equation}
where \(x\) represents an input sampled from \(\mathcal{B}\), \(y_t\) denotes the token generated at time step \(t\), and \(\boldsymbol{y}_{<t}\) denotes the prefix sequence consisting of the first \(t-1\) tokens. In SWERM, a sliding window of size \(k\) is employed to track the average policy entropy over the latest \(k\) training steps, and is defined as:
\begin{equation}
    \bar{\mathcal{H}}_{T}=\frac{1}{k}\sum_{i=T-k+1}^T\mathcal{H}_i
\end{equation}
where \(T\) denotes the current training step, \(\mathcal{H}_i\) denotes the policy entropy at training step \(i\) and \(\bar{\mathcal{H}}_{T}\) is the average entropy computed over the sliding window at step \(T\). To monitor the stability of entropy reduction during training, we quantify the drop in \(\bar{\mathcal{H}}\) from step \(T-1\) to step \(T\) as follows:
\begin{equation}
    \Delta \bar{\mathcal{H}}_T = \bar{\mathcal{H}}_{T-1} - \bar{\mathcal{H}}_T  
\end{equation}
\(\Delta\bar{\mathcal{H}}_T\) serves as an effective indicator for measuring the smoothness of entropy drop. When \(\Delta\bar{\mathcal{H}}_T > \tau\) (where \(\tau\) is a threshold hyperparameter), it indicates a collapse in \(\mathcal{H}_T\), which significantly pulls down \(\Delta\bar{\mathcal{H}}_T\). In contrast, \(\Delta\bar{\mathcal{H}}_T \le \tau\) suggests that the policy entropy is dropping smoothly. Accordingly, to mitigate entropy collapse, we increase the policy temperature and resample the outputs on the current batch whenever \(\Delta\bar{\mathcal{H}}_T > \tau\) is detected.
\section{Prompts Employed in Atom-Searcher}
\begin{figure*}[!t]
\begin{center}
\centerline{\includegraphics[width=1.015\textwidth]{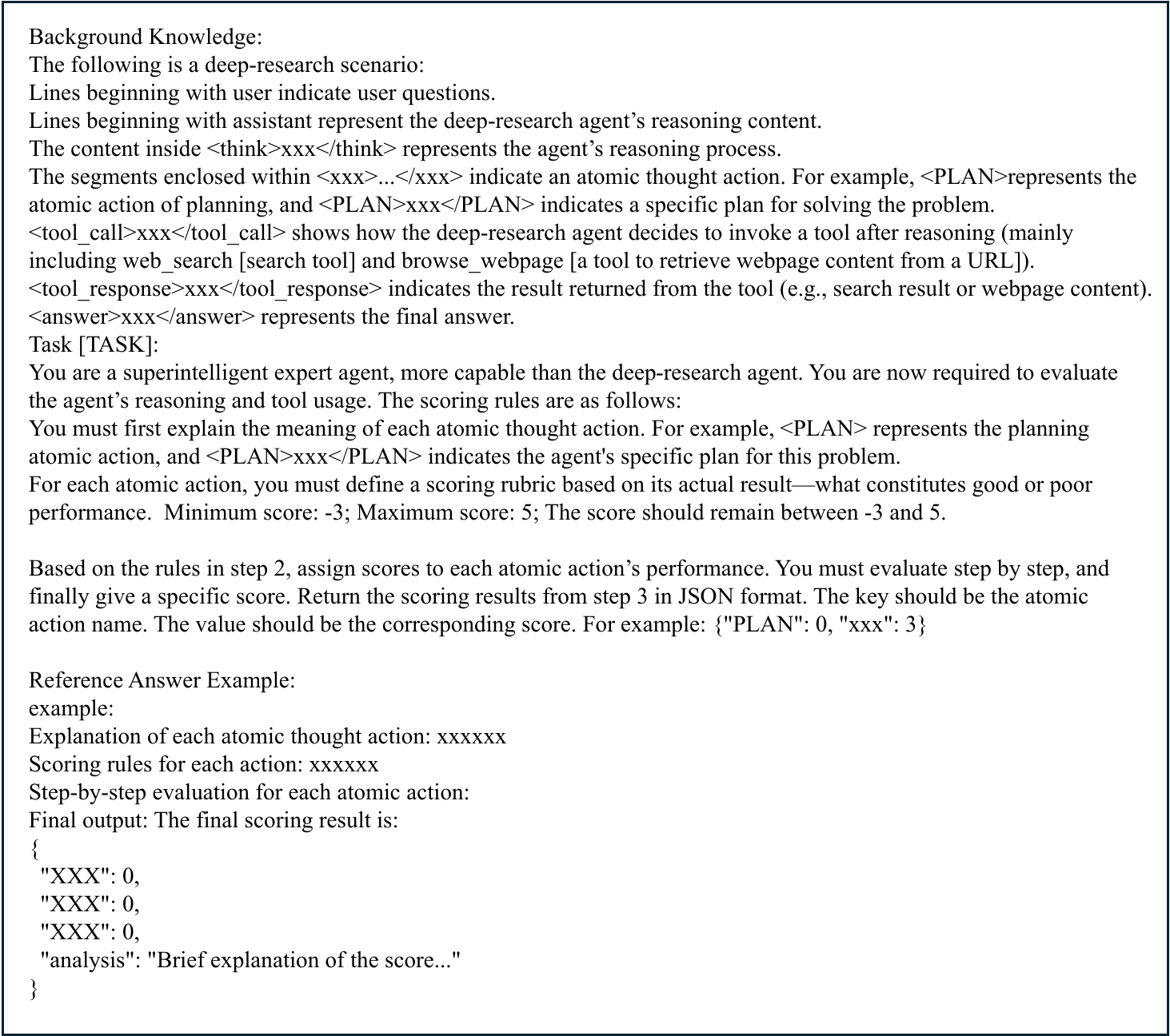}}
\vspace{-2mm}
\caption{
Prompt for RRM to assess the Atomic Thoughts.}
\label{fig:RRM Atom Prompt}
\end{center}
\vskip -0.3in
\end{figure*}

\begin{figure*}[!t]
\begin{center}
\centerline{\includegraphics[width=0.85\textwidth]{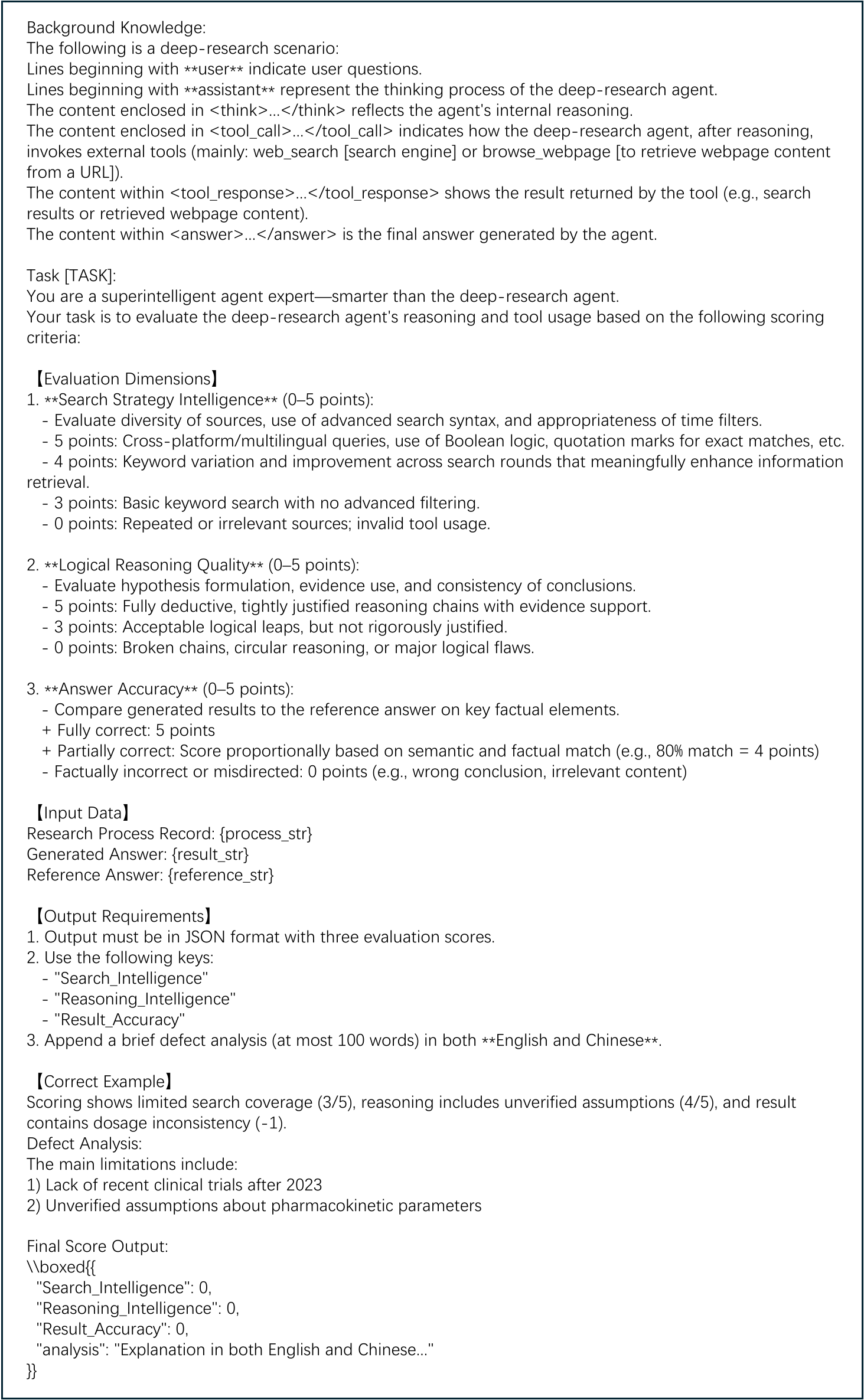}}
\vspace{-2mm}
\caption{
Prompt for RRM to assess the Thought Process.}
\label{fig:RRM Thought Prompt}
\end{center}
\vskip -0.3in
\end{figure*}
\end{document}